\documentclass[journal]{IEEEtran}
\usepackage{cite}

\ifCLASSINFOpdf
   \usepackage[pdftex]{graphicx}
\else
\fi
\usepackage{amsmath}
\usepackage{setspace}

\usepackage{soul} 
\usepackage{booktabs} 

\usepackage[colorlinks=true,linkcolor=blue,citecolor=blue,urlcolor=black ]{hyperref}%

\usepackage{graphicx}
\usepackage{algorithmic}
\usepackage{subfigure}
\usepackage[ruled]{algorithm}

\usepackage[square,numbers]{natbib}
\usepackage{mathtools}

\usepackage{cite}

\begin{document}
%
\title{Local Motion Planner for Autonomous Navigation in Vineyards with a RGB-D Camera-Based Algorithm and Deep Learning Synergy}
%
%
%

\author{Diego~Aghi,
        Vittorio~Mazzia,
        and~Marcello~Chiaberge
\thanks{The authors are with Politecnico di Torino -- Department of Electronics and Telecommunications and Department of Environment, Land and Infrastructure Engineering, Italy. Email: \{name.surname\}@polito.it.}}

%
%

\markboth{}%
{Diego Aghi \MakeLowercase{\textit{et al.}}: Local Motion Planner for Autonomous Navigation in Vineyards with a RGB-D Camera-Based Algorithm and Deep Learning Synergy}
%



\maketitle

\begin{abstract}
With the advent of agriculture 3.0 and 4.0, in view of efficient and sustainable use of resources, researchers are increasingly focusing on the development of innovative smart farming and precision agriculture technologies by introducing automation and robotics into the agricultural processes. Autonomous agricultural field machines have been gaining significant attention from farmers and industries to reduce costs, human workload, and required resources.~Nevertheless,~achieving sufficient autonomous navigation capabilities requires the simultaneous cooperation of different processes; localization, mapping, and path planning are just some of the steps that aim at providing to the machine the right set of skills to operate in semi-structured and unstructured environments. 
In~this~context, this~study presents a low-cost, power-efficient local motion planner for autonomous navigation in vineyards based only on an RGB-D camera, low range hardware, and a dual layer control algorithm.~The first algorithm makes use of the disparity map and its depth representation to generate a proportional control for the robotic platform. Concurrently, a second back-up algorithm, based on representations learning and resilient to illumination variations, can take control of the machine in case of a momentaneous failure of the first block generating high-level motion primitives. Moreover, due to the double nature of the system, after initial training of the deep learning model with an initial dataset, the strict synergy between the two algorithms opens the possibility of exploiting new automatically labeled data, coming from the field, to extend the existing model's knowledge.
The machine learning algorithm has been trained and tested, using transfer learning, with acquired images during different field surveys in the North region of Italy and then optimized for on-device inference with model pruning and quantization. Finally,~the~overall system has been validated with a customized robot platform in the appropriate environment.
\end{abstract}


%
\IEEEpeerreviewmaketitle

\section{Introduction}
\label{intro}
Nowadays, with the continuous growth of the human population, agriculture industries and farmers have been facing the exponential augmentation of global demand of food production. According to the projections of growth established in 2017 by the United Nations~\cite{desa2013world}, by 2050, the global population will be around 9.7 billion and it is expected to reach 11.1 billion in 2100. So, there is an incremental need of new techniques and technologies aimed at maximizing efficiency and productivity of every single land sustainably. 

Over the years, precision agriculture \cite{mulla2013twenty} and digital farming~\cite{r2018research} have gradually contributed with autonomous robotic machines and information collection to improve crop yield and resource management{, to reduce the labor costs, and in part, to increase the production efficiency. This has led to equip harvesting machineries with driverless systems in order to maximize the navigation efficiency by reducing the number of intersections in the path, and therefore, the amount of fuel consumed }\cite{payne2005technologies}{. Once~endowed with the appropriate effectors, these robotic vehicles can harvest }\cite{bac2014harvesting,rath2009robotic}{, spray~}\cite{berenstein2010grape,shapiro2009toward,monta1995agricultural}{, seed }\cite{katupitiya2007systems}{ and irrigate }\cite{kohanbash2012irrigation}{, and collect trees and crops data for inventory management~}\mbox{\cite{virlet2017field,nuske2011yield,wang2013automated}}{; when~configured as platforms, they can carry laborers to prune and thin trees, hence reducing inefficiencies and injuries in the workplace}\cite{davis2012cmu}. Research on applications of mobile robotic systems in agricultural tasks has been increasing vastly~\cite{sharifi2015novel}. However, despite the rising in investments and research activities on the subject, many implementations remain experimental and far from being applied on a large scale. Indeed, most of the proposed solutions require a combination of real-time kinematic GPS (RTK-GPS)~\cite{guzman2016autonomous,dos2016towards} and costly sensors like three-dimensional multi-channel Light Detection and Ranging (LIDAR)~\cite{astolfi2018vineyard}. Other than being very expensive, those solutions are unreliable and prone to failure and malfunction due to their complexity. 

On the other hand, several recent works~\cite{santos2019path,zoto2019automatic} focus their efforts on finding an affordable solution for the generation of a global map with related way-points. However, path following inside vineyard rows is still a challenging task due to localization problems and variability of the environment. Indeed, GPS receivers require to function in an open area with a clear view of the sky~\cite{ma2001gps}, hence, expensive sensors and solutions are needed in order to navigate through vineyards rows and follow the generated~paths.

In this context, we present a low-cost, robust local motion planner for autonomous navigation in vineyards trying to overcome some of the present limitations. Indeed, our power-efficient solution makes use only of RGB-D camera without involving any other expensive sensor such as LIDAR or RTK-GPS receivers. Moreover, we exploit recent advancements in the Deep Learning~\cite{lecun2015deep} techniques and optimization practices for Edge AI~\cite{mazzia2020real} in order to create an overall resilient local navigation algorithm able to navigate inside vineyard rows without any external localization system. 
The machine learning model has been trained using transfer learning with images acquired during different field surveys in the North region of Italy and then we validated the navigation system with a real robot platform in the relevant environment. 

\section{Related Works} 
\label{RelW}
As far as autonomous navigation is concerned, classic autonomous systems capable of navigating a vineyard adopt high-precision RTK-GPS~\cite{ruckelshausen2009bonirob,stoll2000guidance,thuilot2001automatic,ly2015fully} or by the use of laser scanners combined with GPS~\cite{longo2010multifunctional,hansen2011orchard}.~However, the lack of GPS availability due to environmental conditions such as large canopies, the need for prior surveying
of the area, and unreliable connectivity in certain scenarios make GPS-free approaches desirable~\cite{marden2014gps}.
On the other hand, more modern and recent approaches employ different types of sensors usually combined with each other~\cite{astolfi2018vineyard}. For example, Zaidner et al.  introduced a data fusion algorithm for navigation, which optimally fused the localization data from various sensors; GPS, inertial navigation system (INS), visual odometry (VO) and wheel odometry are fused in order to estimate the state and localization of the mobile platform~\cite{zaidner2016novel}.~However,~as~highlighted by the authors, there is a trade-off between cost and accuracy, and the data fusion algorithm could fail if each sensor highly differs from each other. 

Regarding affordable and low-cost solutions, Riggio et al. proposed a low-cost solution based only on a single-channel LIDAR and odometry, but it is greatly affected by the type of canopy and condition of the specific vineyard~\cite{riggio2018low}.~Instead, in~\cite{sharifi2015novel}, they proposed a vision based-control system using a clustering algorithm and Hough Transform in order to detect the central path between two rows. However, it is extremely sensitive to illumination conditions and intra-class variations. 

{On the other hand, the emerging needs of automation in the agricultural production systems and in the crop life cycles, concurrent to the unstoppable expansion towards new horizons of the deep learning, led to the development of several architectures for a variety of applications in precision agriculture.~For instance, in}~\cite{mohanty2016using} {and }\cite{sladojevic2016deep},{ authors proposed solutions based on known architectures such as AlexNet}~\cite{krizhevsky2012imagenet}{, GoogleNet}~\cite{szegedy2015going}{ and CaffeNet}~\cite{jia2014caffe} {to detect diseases in plants and leaves respectively. Moreover, deep learning has been used for crop type classification }\cite{kussul2017deep,mortensen2016semantic,rebetez2016augmenting, mazzia2020improvement}{, crop yield estimation }\cite{kuwata2015estimating,minh2017deep,mazzia2020uav,khaliq2019refining}{, fruit counting }\cite{chen2017counting,bargoti2017deep,mazzia2020real}{, and even to predict the weather, forecasting temperature and rainfall}\cite{sehgal2017crop}.

We propose a novel approach based only on a consumer-grade RGB-D camera and latest innovations in the machine learning field to create a robust to noise local motion planner.~The algorithm that exploits the depth map produces a proportional control and is supported, in case of failure, by a deep learning model that produces high-level primitives~\cite{mousavian2019visual}. Moreover, the strict synergy between the two system blocks opens the possibility to easily create an incremental learning architecture where new labeled data coming from the field are used to extend the existing capabilities of the machine learning model.

The remainder of this paper is organized as follows. Section~\ref{section:sec2} introduces the materials and data used for this research. Sections~\ref{section:se3} and~\ref{section:se4} give a detailed overview of the proposed methodology with the obtained experimental results followed by the conclusion and future works.

\section{Materials and Data}
\label{section:sec2}
In order to acquire a dataset for training and testing the deep neural network, we performed field surveys in two distinct rural areas in the North part of Italy; Grugliasco near the metropolitan city of Turin in the Italian region of Piedmont and Valle San Giorgio di Baone in the Province of Padua in the Italian region Veneto.~The collected video samples present different types of terrains, wine quality, and they were acquired at a different time of the day, with diverse meteorological conditions. Videos were shot at 1080p with a 16:9 ratio in order to have more flexibility during the data processing process.

On the other hand, to acquire images and compute the depth map on the platform, we employed the stereo camera Intel RealSense Depth Camera D435i\footnote{\url{https://www.intelrealsense.com/depth-camera-d435i/}}.
~It is a vision system equipped with an RGB camera and two infrared cameras which computes the depth of each pixel of the acquired frame up to 10 meters.

\begin{figure}[H]
    	\centering
    	\includegraphics[width=0.8\linewidth]{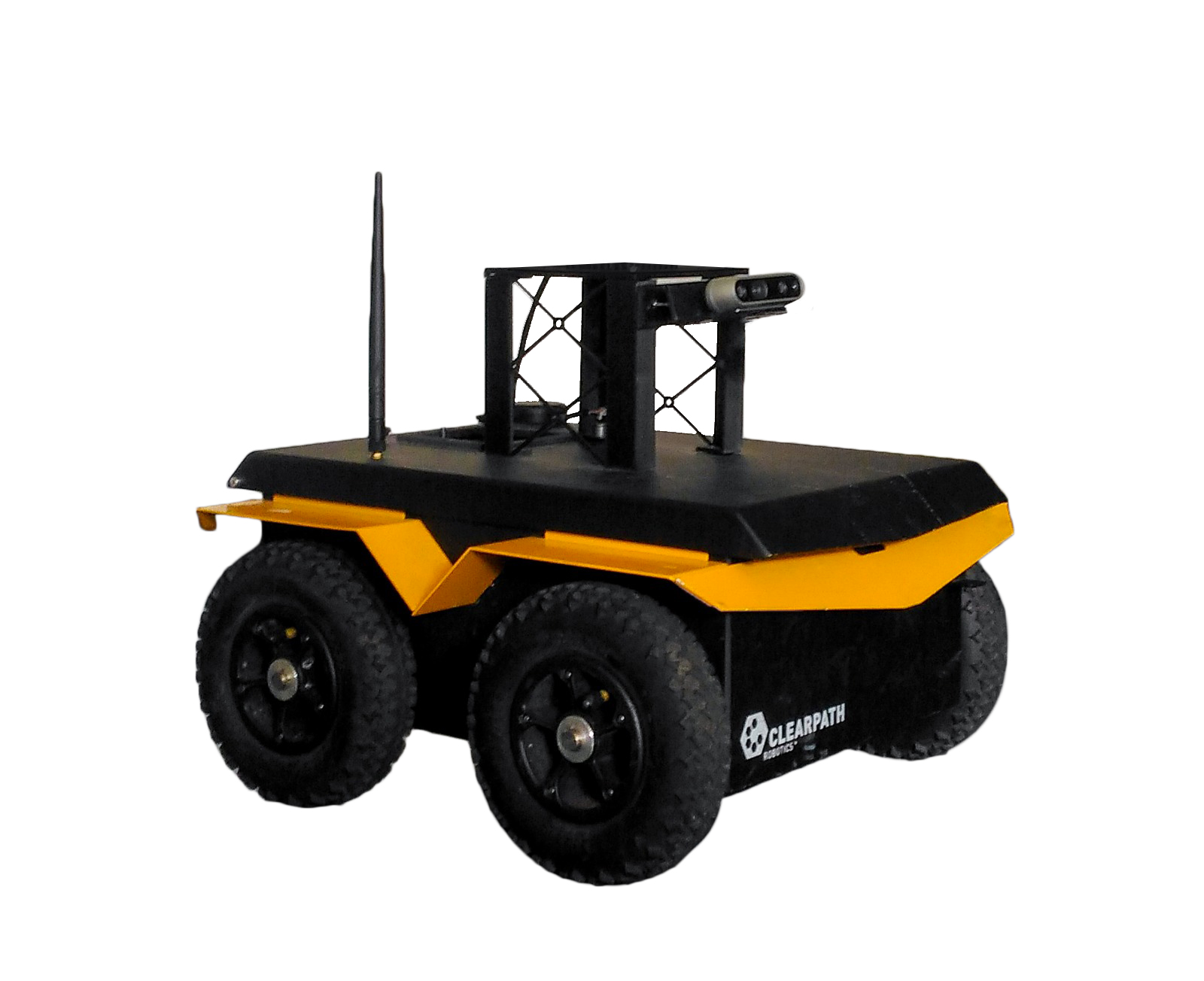}
    	\caption{UGV Jackal from Clearpath Robotics endowed with an Intel RealSense D435i.}
    	\label{fig:jackalrealsense}
    	
    \end{figure}

Finally, for the practical in-field evaluations, the stereo camera has been installed on an unmanned ground vehicle (UGV): the model Jackal from Clearpath Robotic endowed with an Intel Core i3-4330TE.~The camera mounted on the chosen robotic platform is depicted in Figure~\ref{fig:jackalrealsense}. \footnote{\url{https://clearpathrobotics.com/jackal-small-unmanned-ground-vehicle/}} 

\section{Proposed Methodology}
\label{section:se3}
Our goal is to develop a real-time local motion planner with an ultra-light computational load able to overcome practical problems faced by the GPS device when carrying out an autonomous navigation along a vineyard row.

The workflow of our proposal is the following: first, the stereo camera acquires the frames with the RGB camera and, {simultaneously}, it provides a depth map computed through the two infrared cameras. Successively, a light depth{-}map-based algorithm processes the depth maps detecting the end of the vineyard row and, consequently, it calculates the control values with a proportional controller on both linear and angular velocities. Unfortunately, in particular weather and lightning conditions, the~depth map generation is unreliable and prone to error. Indeed, as in many outdoor applications, the~sunlight influences negatively the quality of the results and compromises the control given by the local navigation algorithm. To face this problem, as a back-up solution, we implemented a Convolutional Neural Network (CNN) trained at classifying whether the camera is pointing at the center of the end of the vineyard row or at one of its sides. Once an output prediction is obtained, we can route the path of the robot properly to avoid collisions with the sides of the vineyard. Moreover, we exploited the latest advancement in model optimization techniques in order to obtain an efficient and lightweight neural network able to inference in real-time on a low-cost, low-power device with limited computational capabilities. The overall algorithm pseudo-code is reported in Figure \ref{alg:pseudocode}.
We integrated the proposed algorithm with the open-source Robot Operating System \footnote{\url{https://www.ros.org/}}(ROS) to apply the generated control to the actuators of the selected UGV.
Finally, to prevent the robot platform from colliding with unexpected obstacles that obstruct its way, we use the depth-map provided by the stereo camera {and we apply}  a simple threshold value in order to {immediately} stop the motion in case of an impending collision.

The resulting system is a low-cost, power-efficient, and connection-free local path planner that can be easily integrated with a global system achieving fully autonomous navigation in vineyards.

\subsection{Continuous Depth Map Control}
In order to obtain a proportional control, we detect the center of the end of the vineyard row exploiting the depth-map provided by the stereo camera. Subsequently, the control values for the linear velocity and the angular velocity are calculated proportionally to the horizontal distance between the center of the end of the vineyard row and the longitudinal axis of the UGV.

To this end, we compute the largest area that gathers all the points beyond a certain depth value and then, we bound that area with a rectangle that will be used to compute the control values.

The depth-map is a single-channel matrix with the same dimensions of the image resolution, where each entry represents the depth in millimeters of the corresponding pixel in the camera frame.~The limits of the depth computation are 0 and 8 meters; therefore, the values in the depth matrix range from 0 to 8000.

\begin{figure}[H]
    	\centering
    	\subfigure[]{
\includegraphics[width=0.35\textwidth]{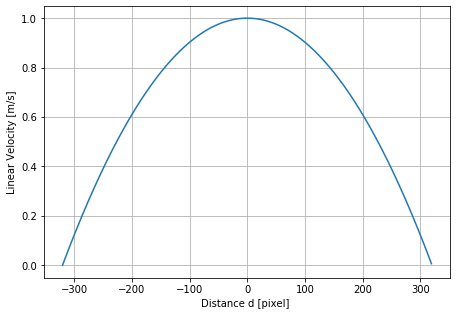}
}
\subfigure[]{
\includegraphics[width=0.35\textwidth]{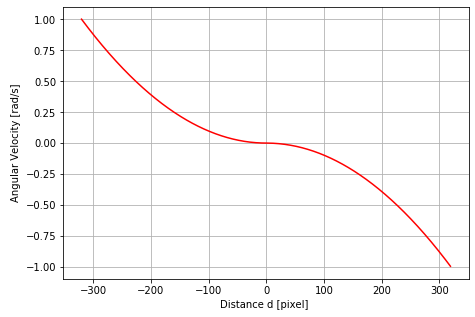}
}
    	\caption{Considering a frame resolution of 640$\times$480, a maximum linear velocity of 1 m/s and a maximum angular velocity of 1 rad/s the plot of the linear velocity control function is shown in Figure~\ref{fig:plotAngVelController}a, whereas, the plot of the angular velocity control function is shown in Figure~\ref{fig:plotAngVelController}b.}
    	\label{fig:plotAngVelController}
    \end{figure}

The main steps of the proposed methodology, described in Algorithm 1, are shown in detailed with the following points:
\begin{enumerate}
    \item \emph{Matrix normalization:} In order to have a solution adaptable to different outdoor scenarios, we need to have a dynamic definition of near field and far-field.~Therefore, we employ a dynamic threshold computed proportionally to the maximum acquired depth value. Hence, by normalizing the matrix, we obtain a threshold that changes dynamically depending on the values of the depth~map.
    \item \emph{Depth threshold: }We apply a threshold on the depth matrix, obtained through a detailed calibration, in order to define which is the near field (represented with a ``0'') and the far field (represented with a ``1''). At this point the depth matrix is a binary mask.
    \item \emph{Bounding operation: }We perform edge detection on the binary image, extrapolating the contours of the white areas, and then, we bound these contours with a rectangle.
    \item \emph{Back-up solution: } If no white area is detected or in case the area of the largest rectangle is less than a certain threshold, we activate the back-up model based on machine learning. 
    \item \emph{Window selection: } On the other hand, if there are multiple detected rectangles, we evaluate only the biggest one in order to get rid of the noise.~The threshold value for the area is obtained through a calibration and it is used to avoid false positive detection. In fact, the holes on the sides of the vineyard row can be detected as large areas with all the points beyond the distance threshold, and therefore they can lead to a wrong command computation. To prevent the system from performing an autonomous navigation using an erroneous detection of the end of the vineyard row, we calibrated the threshold to reduce the possibility that this eventuality occurs drastically. From now on, with the term window we will refer to the largest rectangle detected in the processed frame which area is greater than the area threshold.
    \item \emph{Control values: }The angular velocity and the linear velocity values are both proportional to the horizontal distance (in pixel) between the center of the detected window and the center of the camera frame and based on a parabolic function. 
    The distance \emph{d} is computed as:
    \begin{equation}
    d = X_{w} - X_{c}
    \end{equation} 
    where $X_{w}$ is the horizontal coordinate of the center of the detected rectangle and $X_{c}$ is the horizontal coordinate of the center of the frame. Figure~\ref{fig:schemaaFinale1} shows a graphical representation of the computation of the distance \emph{d}.

   The controller value for the angular velocity ($ang\_vel$) is calculated through the following formula:
    \begin{equation}
    ang\_vel =\begin{cases} - max\_ang\_vel \cdot \left ( \frac{d^{2}}{(\frac{w}{2})^{2}} \right ), & \mbox{if }d\geq 0 \\ max\_ang\_vel \cdot \left ( \frac{d^{2}}{(\frac{w}{2})^{2}} \right ), & \mbox{if }d<0
    \end{cases} 
    \end{equation}
    
where $max\_ang\_vel$ is the maximum angular velocity achievable and \emph{w} is the width of the frame.
    
   \begin{figure}[H]
    	\centering
    	\includegraphics[width=0.86\linewidth]{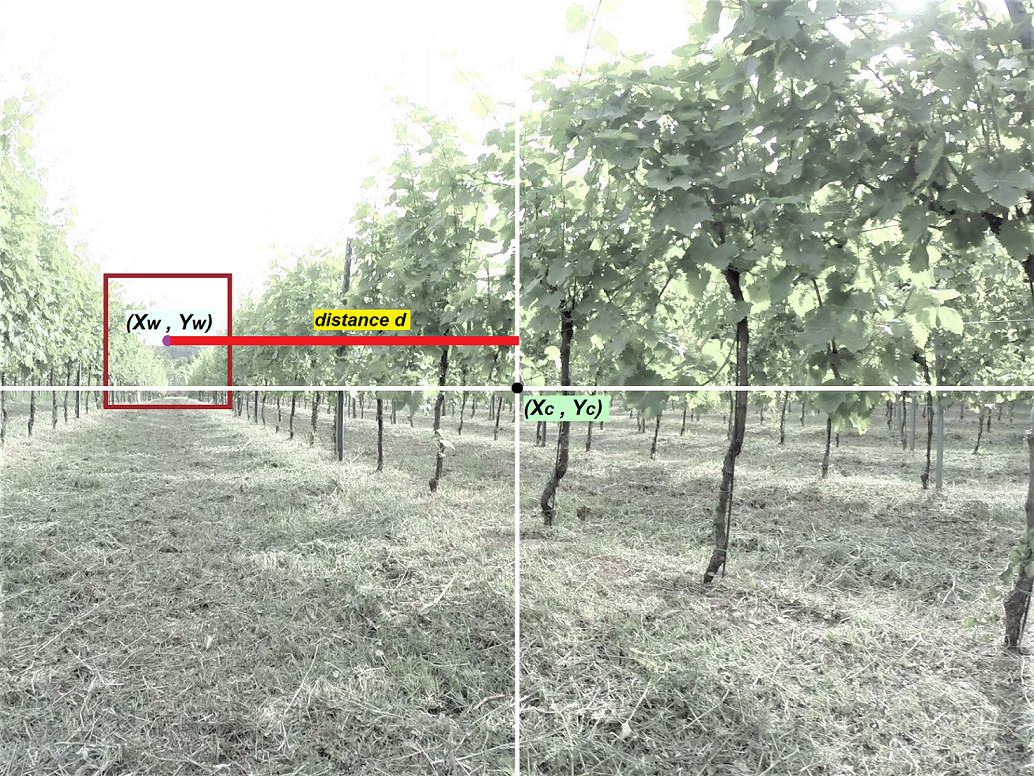}
    	\caption{Depth-map algorithm scheme.~The end of the vineyard row detected is represented by the red rectangle.~The value \emph{d}, on which the control algorithm is based on, is computed as the horizontal distance between the center of the camera frame and the center of the detected window.}
    	\label{fig:schemaaFinale1}
    \end{figure} 
   
\begin{algorithm}[H]
	\caption{RGB-D camera based algorithm}
	\label{alg:pseudocode}
	\begin{spacing}{1.1}

	\begin{algorithmic}[1]
	    \REQUIRE{\textbf{\emph{D}$_{h\times w}$}: Depth Matrix provided by the camera}
	    \REQUIRE{\textbf{F$_{h\times w \times 3}$} RGB frame acquired by the camera}
	    \REQUIRE{T$_{distance}$ threshold on the distance}
	    \REQUIRE{T$_{area}$ threshold on the area of the rectangles}
	    \REQUIRE{$X_c$ horizontal coordinate of the center of the camera frame }
	    \REQUIRE{$X_w$ horizontal coordinate of the center of the detected window}
	    \ENSURE {Control commands for the autonomous navigation}
		\STATE {\textbf{\emph{D}} $\longleftarrow \Vert D \Vert_2$ }
		\FOR {i=1,$\cdots$ h \AND j=1,$\cdots$ w}
		\IF{\textbf{\emph{D}$_{i\times j}$} $>$ T$_{distance}$ }
		\STATE{\textbf{\emph{D}$_{i\times j}$}=1}
		\ELSE
		\STATE{\textbf{\emph{D}$_{i\times j}$}=0}
		\ENDIF
		\ENDFOR
		\STATE{cont[] $\leftarrow$ contours(\textbf{\emph{D}$_{i\times j}$})}
		\STATE{rect[] $\leftarrow$ boundingRect(cont[])}
		\IF{\textbf{max}(area(rect[]))$<$ T$_{area}$ \OR rect[].isEmpty()}
		 \STATE{\textbf{continue} from line 18}
		\ELSE
		   \STATE{angular\_velocity()}
		   \STATE{linear\_velocity()}
		   \STATE{acquire next frame and \textbf{restart} from line 1}
		\ENDIF
		\STATE{\textbf{I$_{1 \times rh \times rw \times 3 }$}$\leftarrow$ preprocessing(\textbf{F$_{h\times w \times 3}$)}}
		\STATE{model\_prediction(\textbf{I$_{1 \times rh\times rw \times 3 }$ })}
		\STATE{ML\_controller()}
		
	\end{algorithmic}
	
	\end{spacing}
\end{algorithm}


    As far as the linear velocity ($lin\_vel$) control function is concerned, it is still be a parabola, but~this time the lower is the distance \emph{d}  the higher its value gets.~Therefore the formula is:
    \begin{equation}
    lin\_vel = max\_lin\_vel \cdot \left ( 1 - \left (\frac{d^{2}}{(\frac{w}{2})^{2}} \right ) \right ) 
    \end{equation}
    where $max\_lin\_vel$ is the maximum linear velocity achievable and \emph{w} is the width of the frame.
    Both control characteristics curve are depicted in Figure~\ref{fig:plotAngVelController}.

\end{enumerate}

\subsection{Discrete CNN Control}
Navigating in an outdoor environment can be extremely challenging. Among several troubles, we noticed that sunlight can be very deceptive when performing edge detection and it could lead to hazardous situations using a total camera-based navigation system.~Therefore, besides the depth-map based algorithm, we propose a back-up solution that exploits machine learning methodologies in order to assist the main algorithm in case of failure.~These are the last points described in Algorithm 1.

 \begin{figure*}[t]
\centering
\includegraphics[width=7.1in]{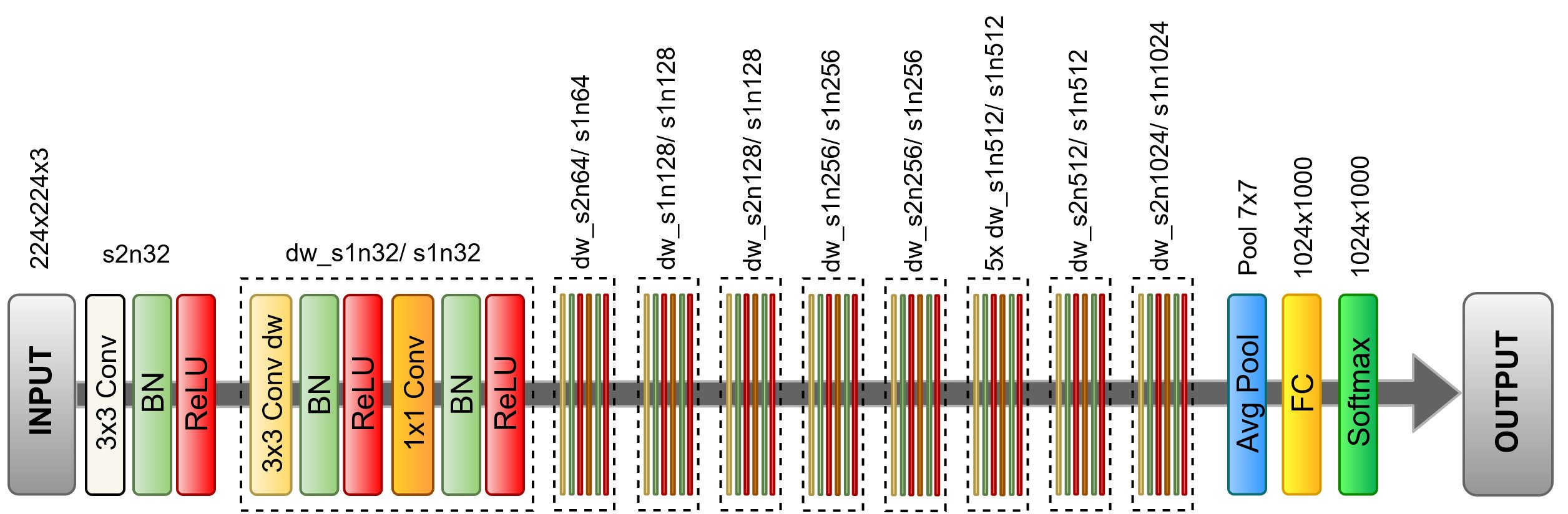}
\vspace{-0.15cm}
\caption{MobileNet network architecture. Depthwise and pointwise convolutions are used through out the entire model drastically reducing the number of parameters required. Coonvolutions with strides $s$ two are used in substitution of Max-Pooling operations.}
\vspace{-0.15cm}
\label{fig:MNArchitecture}
\end{figure*}

Greatly  inspired  by  Giusti  et  al.~\cite{giusti} our second approach relies on a convolutional neural network (CNN) that classifies the frames acquired by the camera into the three following classes: left, center and right. So, in a vineyard scenario, the class center describes the view of the camera when the vehicle is pointing at the end of the vineyard row, whereas the classes left and right indicates whether the vehicle is pointing at the left side or at the right side of the vineyard row, respectively.

Successively, using the predictions of the trained network, we designed a basic control system  to  route  the  path  of  the  robot through  vineyards  rows.  Moreover,  we
exploited latest advancements in model optimization techniques in order to obtain
an efficient and lightweight network able to inference in real-time on a low-cost
edge AI platform.

\subsubsection{Network Architecture}
We have carefully selected a deep learning architecture from the literature that reaches high performance by also containing computational requirements and hardware costs. MobileNet~\cite{mobilenet} network, due to its efficient design, works reasonably fast on mobile devices and embedded systems without too much memory allocation. 

The structure of the MobileNet, illustrated in Figure~\ref{fig:MNArchitecture}, consists on a first convolutional layer with $n=32$ filters and stride $s=2$ followed by 13 layers that include depthwise (dw) and pointwise convolutions. That largely reduces the number of parameters and inference time while still providing reasonable accuracy level.~After each convolution, batch normalization~\cite{ioffe2015batch} and ReLU activation function~\cite{nair2010rectified} are applied.~Every two blocks the number of filters is doubled while reducing the first two dimensions with a stride greater than one.~Finally,~an~average pooling layer resizes the output of the last convolutional block and feeds a fully connected layer with a softmax activation function that produces the final classification predictions. 

We have modified the original final fully connected layers of the MobileNet by substituting them with two fully connected layers of 256 and three neurons, respectively.~The resulting model is a CNN network with an overall depth of 90 layers and with just 3,492,035 parameters.

Moreover, we optimized the network model and we sped up the inference procedure by using the framework provided by NVIDIA TensorRT~\cite{vanholder2016efficient}.

\subsubsection{Pre-Processing}
In the pre-processing phase, before feeding the network, we normalize and resize the images to the expected input dimensions $rh \times rw$ of the model. Indeed, the last two fully connected layers chain the network at a fixed input size of the raw data. More specifically for our modified MobileNet, the~input dimensions are $224\times224$.

\section{Experimental Discussion and Results}
In this section, we discuss the details of the deep learning model training with its dataset generation and evaluation.~Furthermore, we introduce the optimization adjustments applied to the network in order to boost the frequency control to 47.15 Hz.~Finally, we conclude with the experimentation data and results gathered during the field tests.
\label{section:se4}

\subsection{Dataset Creation}
\begin{figure*}[t]
\centering
\subfigure[]{
\includegraphics[width=.3\textwidth]{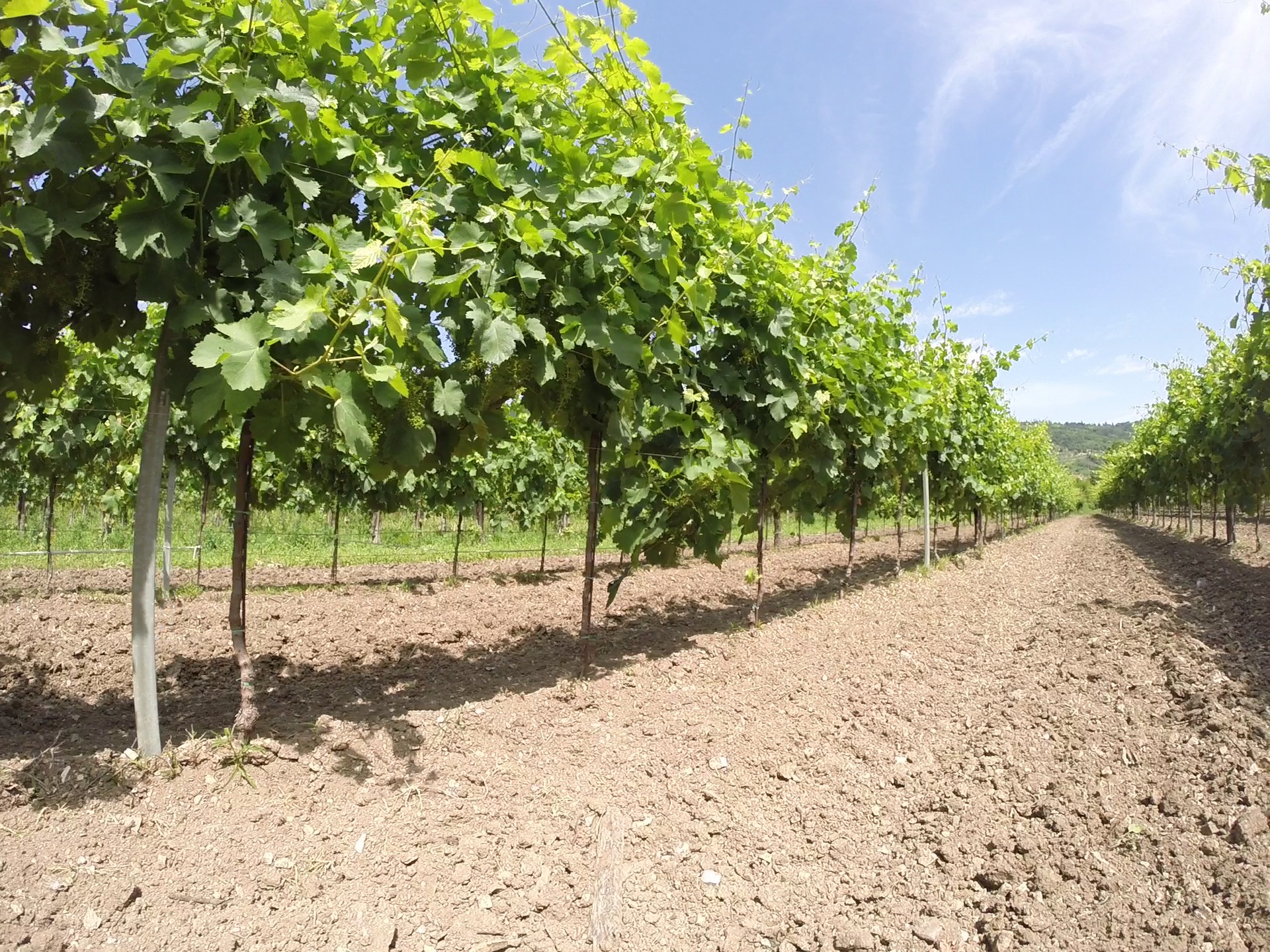}
\label{fig:fig21}
}
\subfigure[]{
\includegraphics[width=.3\textwidth]{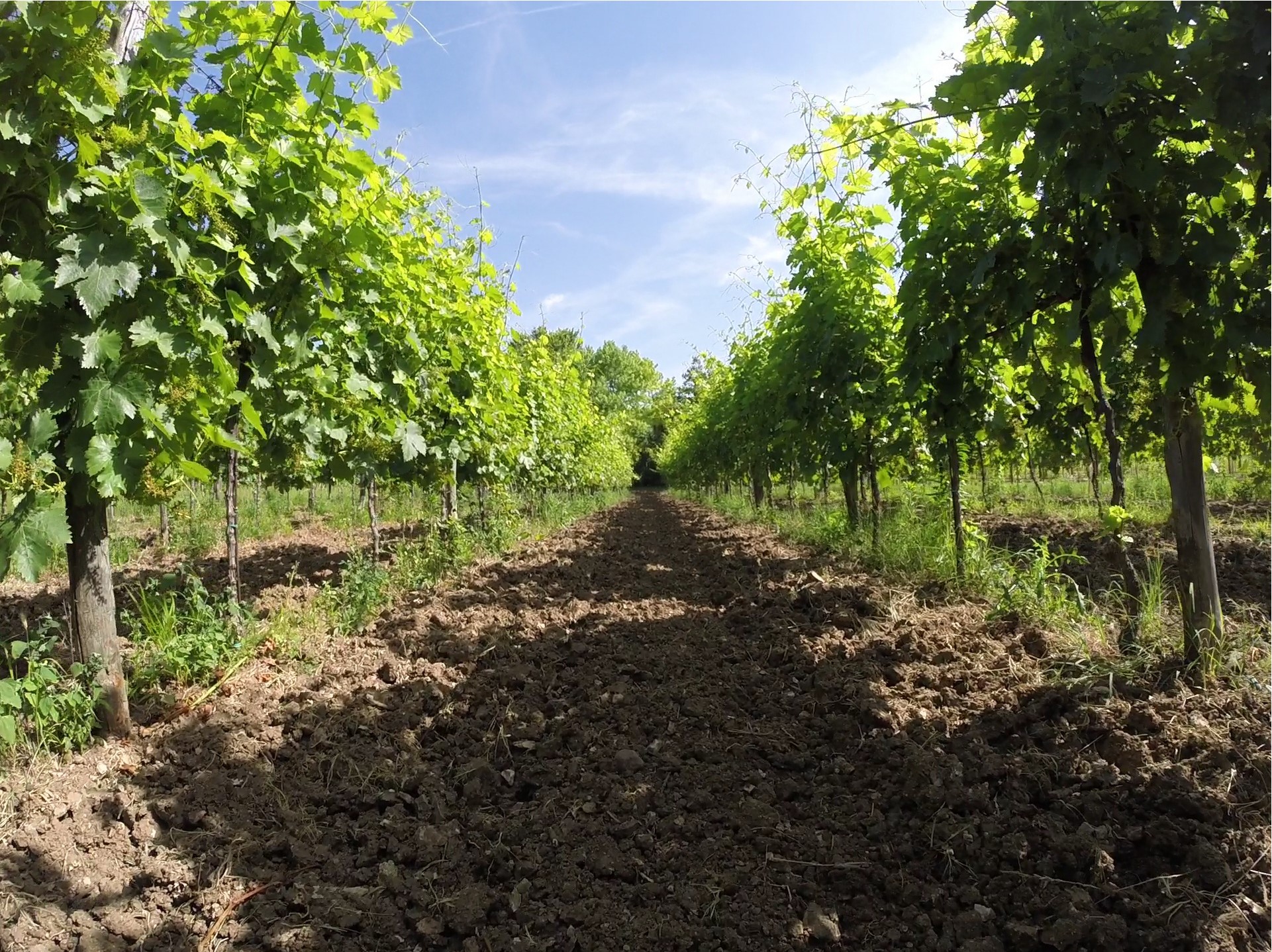}
\label{fig:fig22}
}
\subfigure[]{
\includegraphics[width=.3\textwidth]{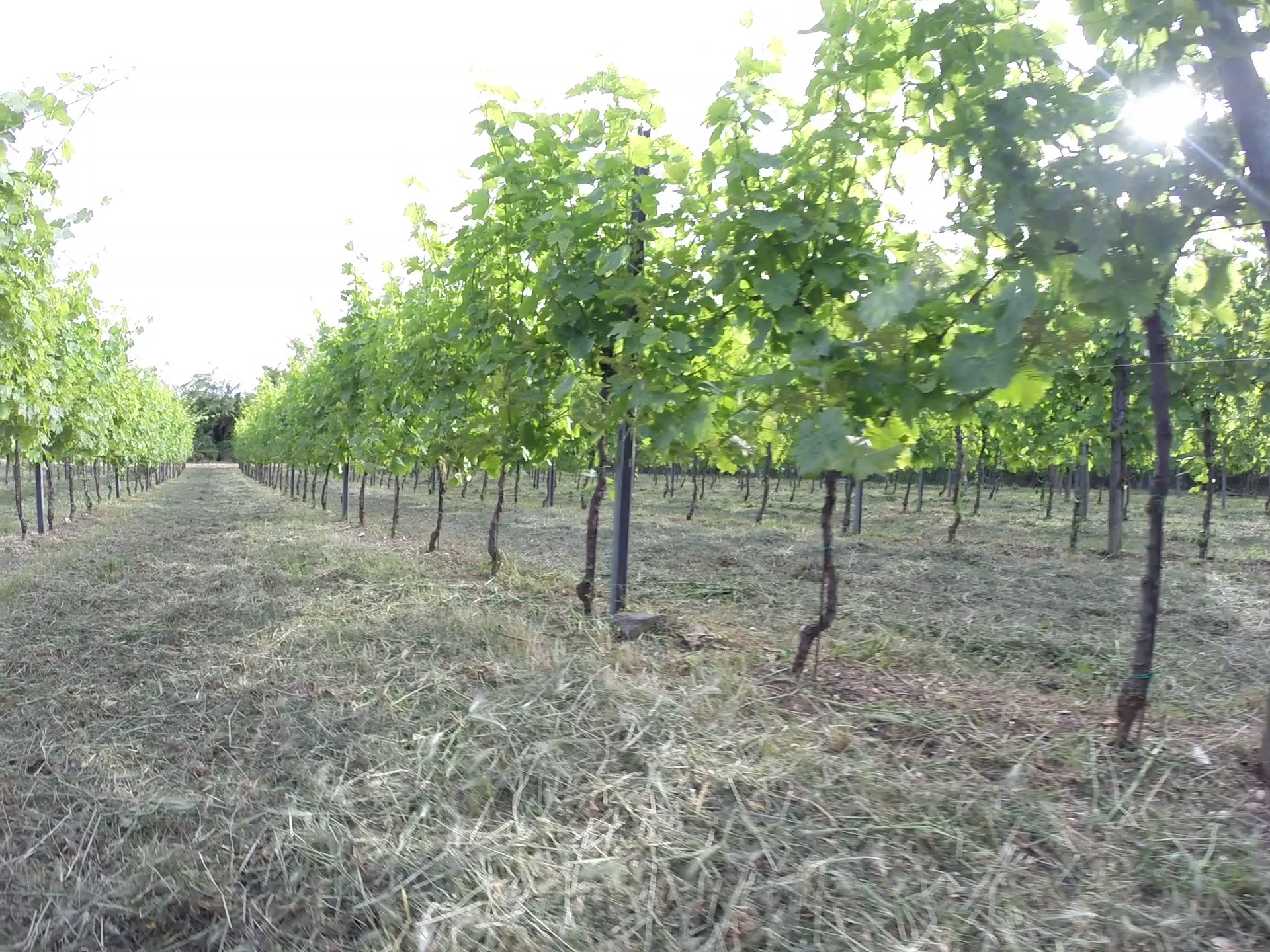}
\label{fig:fig23}
}
\vspace{-0.15cm}
\caption{Three samples of the dataset used to train the network, one for each class. (\textbf{a}) is an example of the left class, (\textbf{b}) of the center class, and (\textbf{c}) of the right class. Dataset samples have been collected with different weather conditions and at a different time of day.~The resulting heterogeneous training set is aimed at giving generality and robustness to the model.}
\vspace{-0.15cm}
\label{fig:datasetSample}
\end{figure*}

We used a dataset of 33.616 images equally balanced along with the three previously introduced classes. In order to create the training dataset, as previously introduced in Section~\ref{section:sec2}, we took several videos in a variety of vineyards rows with a 1080 p resolution camera in order to have more flexibility during the pre-processing phase. In particular, for the first video of the \emph{center} class, we recorded rows with the camera pointing at its center. Whereas, for the other two videos, classes \emph{left} and \emph{right}, we registered with the camera rotated of 45 degrees with respect to the longitudinal axis of the row towards the left and the right side, respectively. Eventually, we took each video as a streaming of images and we selected the best frame every six consecutive ones using a Laplacian filter to detect the less blurring one. Figure~\ref{fig:datasetSample} shows an example for each class.

\subsection{Model Training}
\label{training_section}
As already introduced, we trained the network using a technique known as transfer learning~\cite{tan2018survey}; instead of starting to train with weights randomly initialized, we used variables obtained with an earlier training session.~In particular, we exploited weights obtained fitting MobileNet with the ImageNet classification dataset~\cite{deng2009imagenet}.~Using this technique, we were able to take advantage of previous low-level features, learned by the network, highly reducing the number of images and epochs required for the training. Indeed, edges, contours, and basic textures are general-purpose features that can be reused for different tasks. 
In order to properly train, validate and test the model, we randomly divided the dataset into three subsets as follow: 70\% for the training set, 15\% for the development set, and the remaining 15\% for the test set. We trained the resulting network for only six epochs with a batch size of 64. To increment the robustness of the network and to overcome possible problems of overfitting, we used different techniques such as dropout~\cite{srivastava2014dropout}, weight decay~\cite{loshchilov2018fixing} and data augmentation with changes in zoom and brightness~\cite{perez2017effectiveness}.~Finally, we used mini-batches with RMSprop optimizer~\cite{tieleman2012lecture}, accuracy metric, and cross entropy as a loss function.

\subsection{Machine Learning Model Evaluation and Optimization}
The implemented model has been trained and tested with the subdivision of the dataset introduced in previous Section~\ref{training_section}, giving an accuracy of 1.0 over the test set.~Therefore, this model is the one employed for the navigation.

\begin{figure*}[t]
\centering
\subfigure[]{
\includegraphics[width=.25\textwidth]{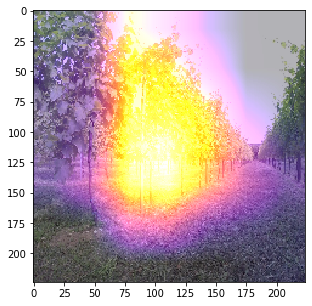}
\label{fig:fig24}
}
\subfigure[]{
\includegraphics[width=.25\textwidth]{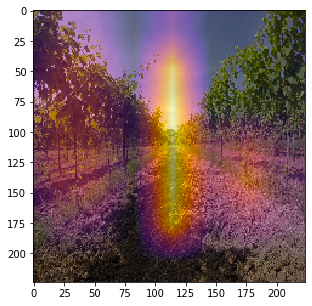}
\label{fig:fig25}
}
\subfigure[]{
\includegraphics[width=.25\textwidth]{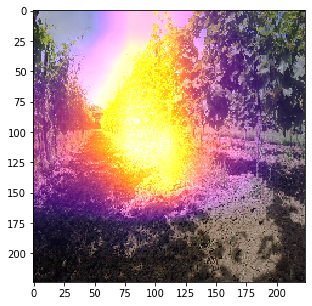}
\label{fig:fig26}
}
\vspace{-0.15cm}
\caption{{Gradient information, flowing into the last convolutional layer of MobileNet, is used to understand each neuron for a decision of interest. It is possible to assess that for either the three classes, (\textbf{a}--\textbf{c}), the network is "looking" at the vineyard roads and their vanishing point.}}
\vspace{-0.15cm}
\label{fig:gradcam}
\end{figure*}

\begin{figure}[H]
    	\centering
    	\includegraphics[width=0.9\linewidth]{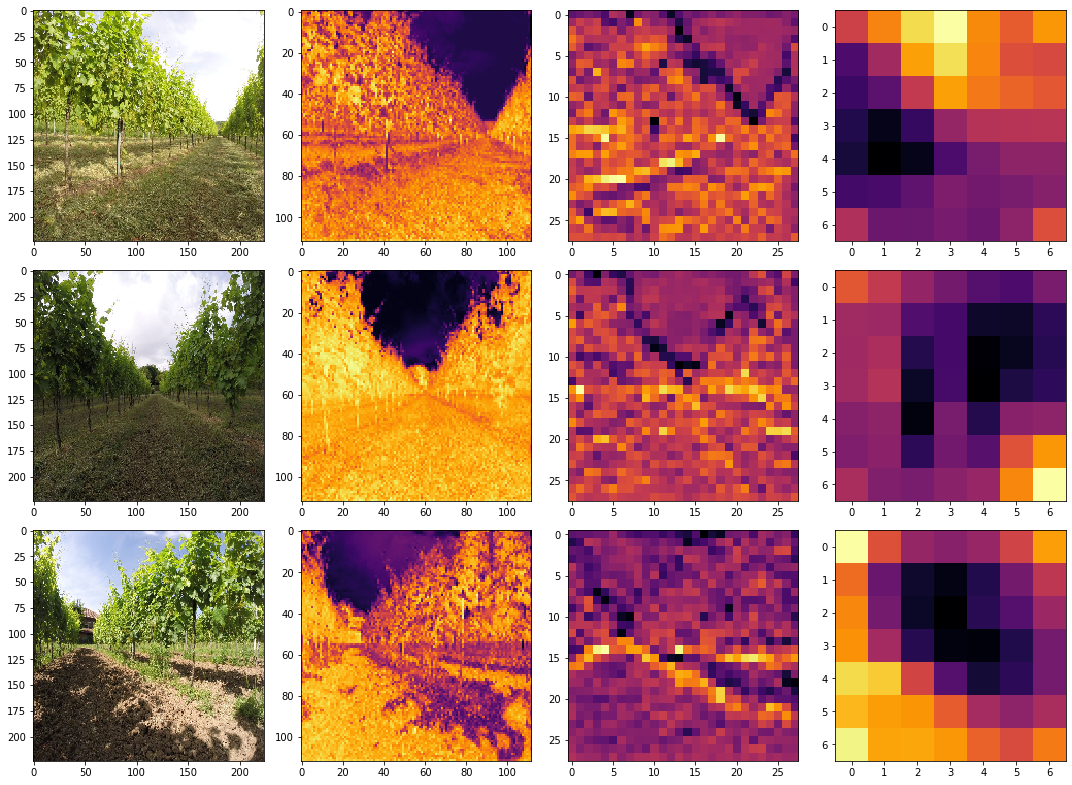}
    	\caption{Three input images belonging to different classes, with their respective activation maps taken at different level of depth of the network. Already on early stages, the network, pre-trained on ImageNet, is able to extract useful representations that lead at robust, disentangle activations in the final layers. It is possible to notice how the two spacial dimensions are increasingly reduced.}
    	\label{fig:predictions_layers}
\end{figure}

In order to inspect the model and justify the high accuracy of it, we plotted the intermediate activations of the trained network {and we adopted Grad-CAM,}~\cite{selvaraju2017grad},{ to highlight important regions in the image for predicting the correct class}. In Figure~\ref{fig:predictions_layers} are shown some feature maps at different level of depth: immediately after the first convolution, at an intermediate point and before the average pooling layer. It is clear how the deep learning model is able to generate robust feature maps already after the first convolution. Later those representations are exploited in order to produce disentangled representations that easily allow the model to predict the three different classes with high level of confidence. 
{Instead, in Figure}~\ref{fig:gradcam}, {are presented the regions of interest for the three different classes. With Grad-CAM we can visually validate that the network is activating around the proper patterns of the input image and that it is not exploiting short cuts to achieve a high level of accuracy. Indeed, we can easily assess that the model, trained with transfer learning, is exploiting the vineyard rows and their vanishing point to obtain an effective generalization power.}

Moreover, in order to evaluate the robustness of the network over new scenarios, and prove how transfer learning is so effective for this specific application, we performed an experimentation, training the model only with a small part of the available dataset.~So, we trained the architecture with just a vineyard type and tested the resulting model with five completely different scenarios with diverse wine quality and weather conditions. In particular, we used only 18\% as training examples, corresponding to 6.068 images, due to the amount of images available for each region of the available dataset. Consequently, we tested the new trained network with the remaining 27458 samples. 

As shown in Table~\ref{tab:tab2}, an accuracy of 0.94 is achieved by the re-trained model in this second case. That is an optimal result considering the fact that the network has been trained with a very small dataset and it has been tested with a completely different vineyard scenario. This clearly demonstrates how transfer learning, for this specific task, is very effective at providing good generalization capabilities with also a small training set. {We also compared, using this last split, the selected network with other notable architectures of the literature. As it is clear from Table}~\ref{tab:tab3} {MobileNet is the right balance between average accuracy and computational request. However, in the presence of a platform with more flexible computational constraints, EfficientNet-B1, or networks with higher compound coefficient $\phi$}~\cite{tan2019efficientnet}, {would be much more likely to generalize over new scenarios maintaining an optimal level of~efficiency.} 

\begin{table}[H]
\begin{center}
\caption{Results of the second evaluation of the trained CNN model.}
\label{tab:tab2}
\begin{tabular}{cccc}
\toprule
\multicolumn{1}{c}{\textbf{Class}}                       & \multicolumn{1}{c}{\textbf{Precision}} & \multicolumn{1}{c}{\textbf{Recall}} & \multicolumn{1}{c}{\textbf{f1-Score}} \\ \midrule
\multicolumn{1}{c}{\textbf{right}}        & \multicolumn{1}{c}{0.850}              & 
\multicolumn{1}{c}{1.000}           & \multicolumn{1}{c}{0.919}             \\ 
\multicolumn{1}{c}{\textbf{left}}         & \multicolumn{1}{c}{1.000}              & \multicolumn{1}{c}{0.899}           & \multicolumn{1}{c}{0.947}             \\ 
\multicolumn{1}{c}{\textbf{center}}       & \multicolumn{1}{c}{1.000}              & \multicolumn{1}{c}{0.924}           & \multicolumn{1}{c}{0.961}             \\  \midrule
\multicolumn{1}{c}{\textbf{micro avg}}    & \multicolumn{1}{c}{0.941}              & \multicolumn{1}{c}{0.941}           & \multicolumn{1}{c}{0.941}             \\ \midrule
\multicolumn{1}{c}{\textbf{macro avg}}    & \multicolumn{1}{c}{0.950}              & \multicolumn{1}{c}{0.941}           & \multicolumn{1}{c}{0.942}             \\ \midrule
\multicolumn{1}{c}{\textbf{weighted avg}} & \multicolumn{1}{c}{0.950}              & \multicolumn{1}{c}{0.941}           & \multicolumn{1}{c}{0.942}             \\ 
\bottomrule
\end{tabular}
\end{center}
\end{table}
Finally, as previously introduced, the employed network has been optimized, discarding all redundant operations and reducing the floating point precision from 32 to 16 bits, using the framework TensorRT.~The optimization process, besides not affecting the accuracy of the predictions, it gives a significant increment to the number of frames elaborated per second by our model, using the same hardware supplied with the robot. In fact, the control frequency using Tensorflow with a frozen graph, computational graph of the network without optimization and tranining nodes, was 21.92 Hz, whereas, with the performed optimization, we reached 47.15 Hz.

\begin{table}[H]
\begin{center}
\caption{{Comparison between different CNN architectures using 18\% of the available dataset and transfer learning to train the networks. MobileNet, for this specific application, is a good trade-off between performance and computational cost.}}
\label{tab:tab3}
\begin{tabular}{llll}
\toprule
\textbf{Model}  & \textbf{Parameters} & \textbf{GFLOPSs} & \textbf{Avg. Acc.} \\ \midrule
MobileNet~\cite{mobilenet}       & 4,253,864           & 0.579            & 94.7\%             \\ 
MobileNetV2~\cite{sandler2018mobilenetv2}    & 3,538,984           & 0.31             & 91.3\%             \\
EfficientNet-B0~\cite{tan2019efficientnet} & 5,330,571           & 0.39             & 93.8\%             \\ 
EfficientNet-B1~\cite{tan2019efficientnet} & 7,856,239           & 0.70             & 96.8\%             \\ 
ResNet50~\cite{he2016deep}       & 25,636,712          & 4.0              & 93.1\%             \\ 
DenseNet121~\cite{huang2017densely}   & 8,062,504           & 3.0              & 95.3\%             \\ \bottomrule
\end{tabular}
\end{center}
\end{table}

\subsection{Field Experimentation}
\begin{figure}[H]
    	\centering
    	\includegraphics[width=0.65\linewidth]{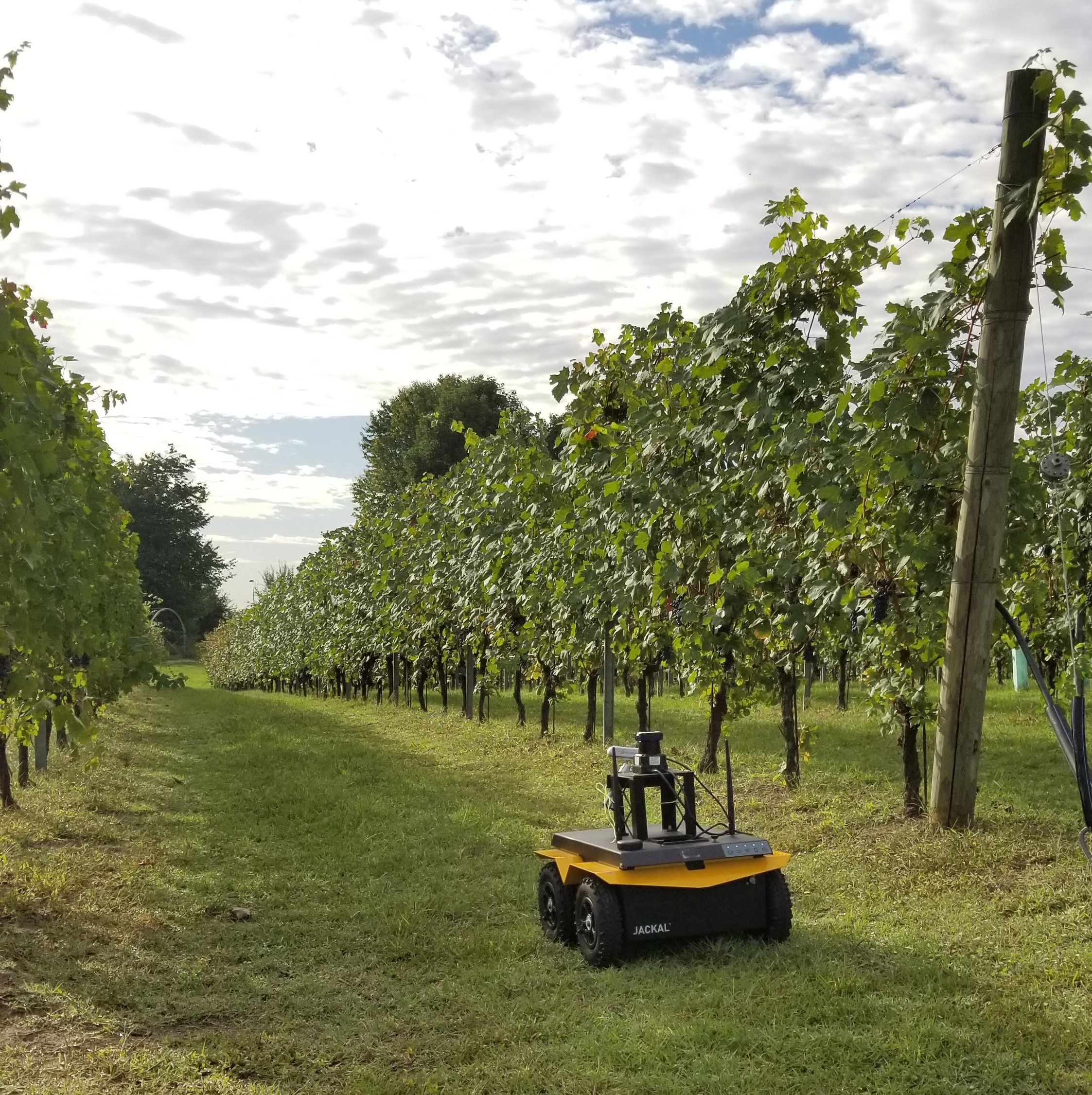}
    	\caption{Jackal model robotic platform on the test site during the field experimentation.}
    	\label{fig:jackal}
    \end{figure}

As far as the deployment is concerned, the system has been implemented in a ROS-oriented robot platform. The robot in which the local planner has been tested is an unmanned ground vehicle: the model Jackal from Clearpath Robotics (Figure \ref{fig:jackal}) introduced in Section~\ref{section:sec2}.

The tests have been carried out in a new vineyard scenario. In order to correctly perform navigation the stereo camera has been installed in such a way that the center of the camera frame corresponds to the longitudinal axis of the vehicle. 

 The proposed solution, after several trials with different vineyard rows {but similar weather conditions}, proved to be able to perform an autonomous navigation along the given paths, even~lowering down the resolution of the camera to $640\times480$. {More specifically, for the two infrared cameras with which the camera computes the depth-map the resolution has been set to $640\times480$, whereas, for the RGB images processed by the machine learning model we started with a resolution of $1280\times720$ and then we gradually reduced it until $640\times480$ as mentioned. Moreover, when acquiring images we used the default calibration provided by Intel with the camera lens distortion based on the Brown--Conrady model}~\cite{duane1971close}{.~The intrinsic parameters for the final configuration of the camera in both the so-called depth and color modes are showed in Table}~\ref{tab:tabparam}. All our tests showed precise trajectories, comparable with ones obtained with data fusion techniques that make use of several expensive sensors to maintain the correct course.

\begin{table}[H]
\begin{center}
\caption{{Intrinsic parameters of the camera for each of the exploited modes. Principal point (PPX and PPY) and focal length (Fx and Fy) for a resolution of $640\times480$.}}
\label{tab:tabparam}
\begin{tabular}{lcc}
\toprule
\textbf{Mode} & \textbf{Depth}            & \textbf{Color}            \\ \midrule
\textbf{PPX}  & 321.910675048828 & 316.722351074219 \\
\textbf{PPY}  & 236.759078979492 & 244.21875        \\
\textbf{Fx}   & 387.342498779297 & 617.42242431640  \\
\textbf{Fy}   & 387.342498779297 & 617.789978027344  \\ 
\bottomrule
\end{tabular}
\end{center}
\end{table}
 
\begin{figure}[H]
    	\centering
    	\includegraphics[width=0.9\linewidth]{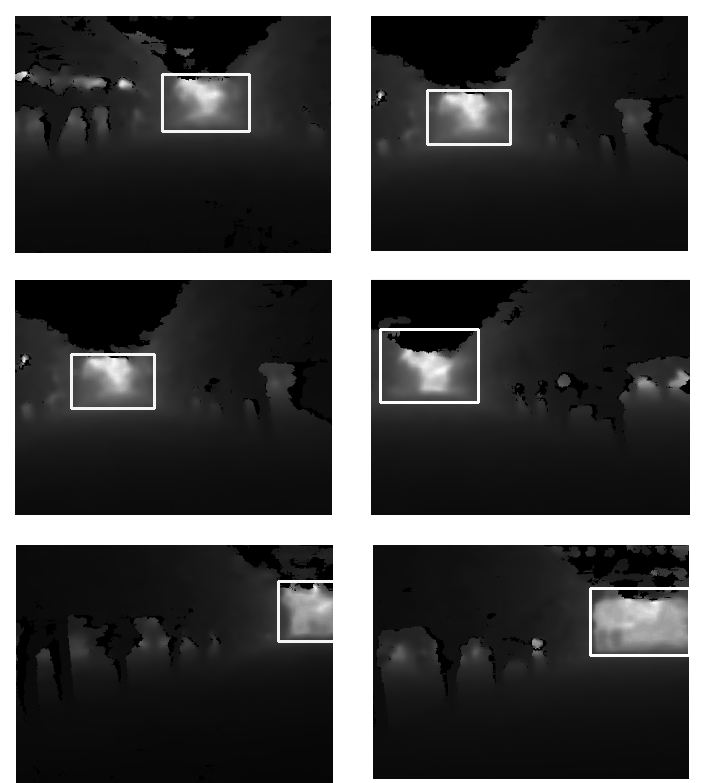}
    	\caption{{Instances of the depth-map based algorithm while performing tests in the vineyards.}  Wherever the robot is pointing at, it is capable of  correctly detecting the end of the vineyard rows. }
    	\label{fig:example2}
    \end{figure}

 As noticeable from the depth maps samples in Figure~\ref{fig:example2} taken during the field experimentation, the first method can detect the end of the vineyard independently from the direction of the longitudinal axis of the robot.~The rectangle is successively used for control signals generation. In case of a fault of this solution, as previously introduced, the machine learning based algorithm takes control. Finally,~it~is possible to exploit the distance value \emph{d} to easily collect new, already labeled, sample data from the operational work of the robotic platform. Indeed, due to the nature of all mini-batch gradient descent based optimizer, it is possible to continuously use new data points to extend the existing model's knowledge obtaining a more robust and prone to generalize neural network.

\section{Conclusions}

We proposed a local motion planner for vineyards rows autonomous navigation. We exploited the stereo vision properties of an RGB-D camera and latest advancements in deep learning optimization techniques in order to obtain a lightweight, power-efficient algorithm able to run on a low-cost hardware.

The proposed overall methodology provides a real-time control frequency using only hardware with limited computational capabilities containing costs and required resources.~The back-up trained neural network is robust to different factors of variation, and after the optimization procedure, it~provides a control frequency of 47.15 Hz without the need of external hardware accelerators.

Finally, the proposed local motion planner has been implemented on a robotic platform and tested on the relevant environment, demonstrating to scale real working conditions even with a low~resolution. 

As future work, we plan to integrate the presented work with a concrete application and extent the methodology to orchards and any other analogous scenario.

\section*{acknowledgments}
This work has been developed with the contribution of the Politecnico di Torino Interdepartmental Centre for Service Robotics PIC4SeR (https://pic4ser.polito.it) and SmartData@Polito (\url{https://smartdata.polito.it}).

\ifCLASSOPTIONcaptionsoff
  \newpage
\fi


\vfill\break

\begin{IEEEbiography}[{\includegraphics[width=1in,height=1.25in,clip,keepaspectratio]{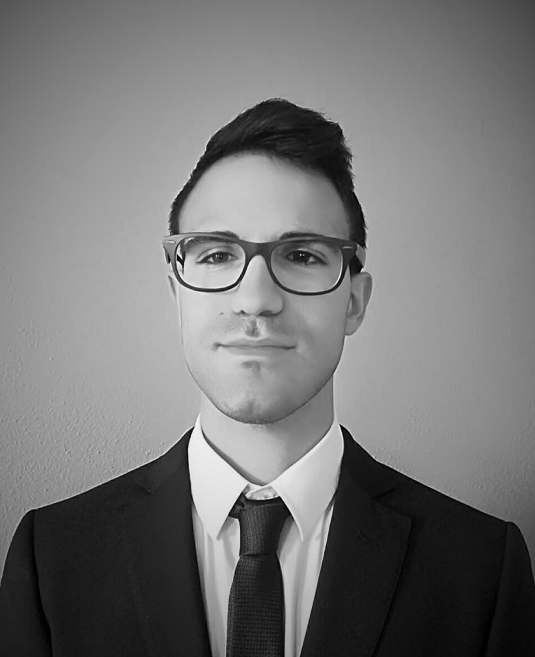}}]{Diego Aghi} is a researcher at PIC4SeR - PoliTO Interdepartmental Centre for Service Robotics (\url{https://pic4ser.polito.it/}).
He graduated from Politecnico di Torino with the thesis “Navigation Algorithms for Unmanned Ground Vehicles in Precision Agriculture Applications” carried out at PIC4SeR. He is now focusing his research activity on the development of machine learning and computer vision algorithms for autonomous navigation applications in outdoor environment.
\end{IEEEbiography}

\begin{IEEEbiography}[{\includegraphics[width=1in,height=1.25in,clip,keepaspectratio]{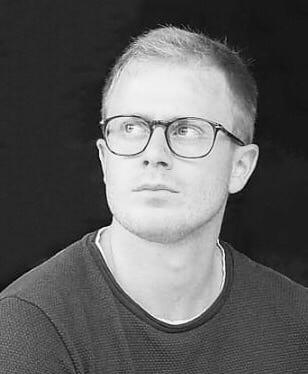}}]{Vittorio Mazzia} is a PhD student in Electrical, Electronics and Communications Engineering working with the two Interdepartmental Centres PIC4SeR (\url{https://pic4ser.polito.it/}) and SmartData (\url{https://smartdata.polito.it/}). He received a master's degree in Mechatronics Engineering from the Politecnico di Torino, presenting a thesis entitled "Use of deep learning for automatic low-cost detection of cracks in tunnels," developed in collaboration with the California State University. His current research interests involve deep learning applied to different tasks of computer vision, autonomous navigation for service robotics, and reinforcement learning. Moreover, making use of neural compute devices (like Jetson Xavier, Jetson Nano, Movidius Neural Stick) for hardware acceleration, he is currently working on machine learning algorithms and their embedded implementation for AI at the edge.
\end{IEEEbiography}

\begin{IEEEbiography}[{\includegraphics[width=1in,height=1.25in,clip,keepaspectratio]{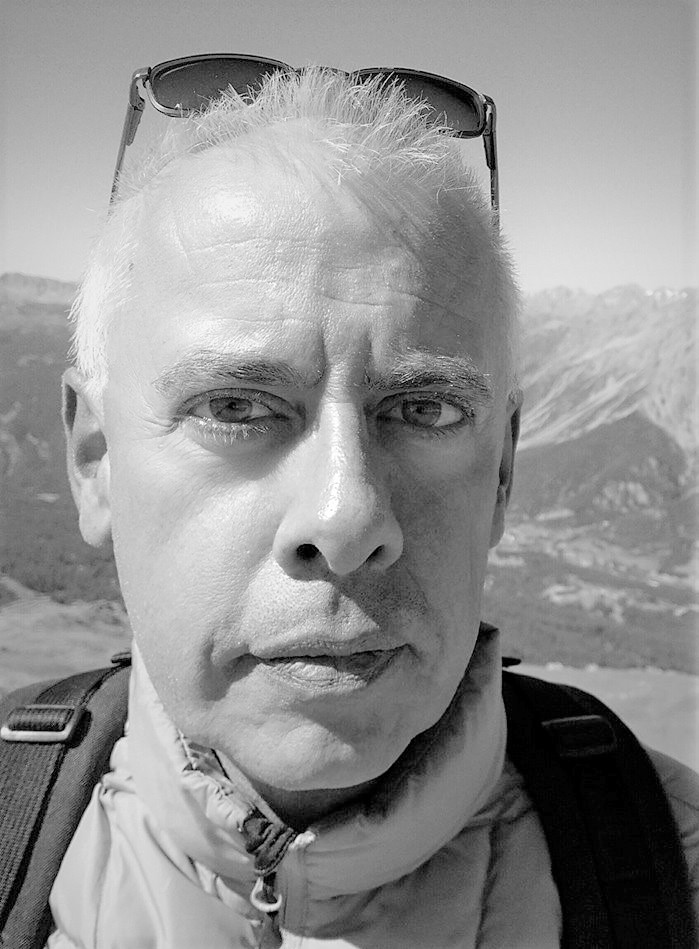}}]{Marcello Chiaberge} is currently Associate Professor within the Department of Electronics and Telecommunications, Politecnico di Torino, Turin, Italy. He is also the Co-Director of the Mechatronics Lab, Politecnico di Torino
(\url{www.lim.polito.it}), Turin, and the Director and the Principal Investigator of the new Centre for Service Robotics (PIC4SeR, \url{https://pic4ser.polito.it/}), Turin. He has authored more than 100 articles accepted in international conferences and journals,
and he is the coauthor of nine international patents. His research interests include
hardware implementation of neural networks and fuzzy systems and the design and implementation of reconfigurable real-time computing architectures. \end{IEEEbiography}
\vfill

\end{document}